\documentclass[runningheads]{llncs}
\usepackage[T1]{fontenc} 
\usepackage{graphicx} 

\newcommand{\code}{\texttt}

\begin{document}
\title{Can Large Language Models put 2 and 2 together? Probing for Entailed Arithmetical Relationships}
\titlerunning{Can LLMs put 2 and 2 together? Probing EAR}

\author{Dagmara Panas\orcidID{0009-0007-7805-2141} \and
Sohan Seth\orcidID{0000-0002-7464-4752} \and
Vaishak Belle\orcidID{0000-0001-5573-8465}}
\authorrunning{D. Panas et al.}

\institute{School of Informatics, The University of Edinburgh, Edinburgh, UK}

\maketitle 
\begin{abstract}
Two major areas of interest in the era of Large Language Models regard questions of what do LLMs \emph{know}, and if and how they may be able to \emph{reason} (or rather, approximately reason). Since to date these lines of work progressed largely in parallel (with notable exceptions), we are interested in investigating the intersection: probing for reasoning about the implicitly-held knowledge. Suspecting the performance to be lacking in this area, we use a very simple set-up of comparisons between cardinalities associated with elements of various subjects (e.g. the number of legs a bird has versus the number of wheels on a tricycle). We empirically demonstrate that although LLMs make steady progress in knowledge acquisition and (pseudo)reasoning with each new GPT release, their capabilities are limited to statistical inference only. It is difficult to argue that pure statistical learning can cope with the combinatorial explosion inherent in many commonsense reasoning tasks, especially once arithmetical notions are involved. Further, we argue that \textit{bigger is not always better} and chasing purely statistical improvements is flawed at the core, since it only exacerbates the dangerous conflation of the production of correct answers with genuine reasoning ability.  

\end{abstract}

\section{Introduction}
The introduction of the Transformer architecture \cite{vaswani2017attention}, together with the unprecedented abundance of data (in)untarily donated by the masses of social media users and digital content producers, has brought about a fundamental shift in Natural Language Processing, which had until then substantially lagged behind Computer Vision. 
Although early forays into straightforward network approaches to analysing textual data yielded a surprising success\footnote{As well as a degree of controversy around the implicit biases in the data exposed through vector algebra.} with the introduction of \emph{word embeddings} \cite{mikolov2013efficient}, arguably it was not until the release of GPT and its successors that questions of knowledge, reasoning abilities, and even sentience \cite{vallance2022bbc} of computers have become a mainstream topic of debate.
Further additions of efficient fine-tuning solutions, as well as code-less options such as in-context few-shot learning or chain-of-thought (CoT) prompting techniques, have enabled these tools to become incredibly useful personal assistants and further contributed to the perception that we are approaching Artificial General Intelligence (AGI) \cite{bubeck2023sparks}.
Juxtaposing the extremely convincing nature of LLMs and great appetite for adoption in every area of society with at times spectacular failure modes \cite{rumbelow2023glitch,li2024glitch} understandably elicits a host of safety, reliability and trustworthiness concerns ~\cite{geiping2024coercing,guo2023evaluating}.
Specifically, while it may often \emph{seem} that a model is a thinking rational agent, not only capable of carrying out a conversation but also performing useful tasks \cite{schick2024toolformer} and a degree of logical inference \cite{pan2023logic} or calculations \cite{bao2024llms}, in reality it is simply performing \textit{(data-)informed guesses}.
Perhaps the most stark illustration of this nature comes from research into memorisation versus stability of acquired knowledge: the first arc shows that an astonishing proportion of data examples can be recovered from an LLM by simply feeding it part of the example \cite{Al_Kaswan_2024,yu2023assessing}; the second arc on the other hand demonstrates that small perturbations in prompts lead to a drastic drop in performance \cite{lin2020birds,jin2023can}.

While there is no doubt that LLMs and their statistically-informed guesses can be of great assistance in many fault-tolerant domains, in areas such as medicine, governance or finance the bar is higher than current capabilities would allow.
There are two ways to try and meet such reliability demands: more data covering more bases \cite{lin2020birds,geva2020injecting,mostafazadeh2020glucose}; or a Neuro-Symbolic (NeSy) approach, be it through hybrid `enablement' with external solvers \cite{pan2023logic} or via integrated architectural solutions \cite{badreddine2022logic,teso2017structured}.
We argue that the first path is doomed in the long run, as it addresses the symptoms, not the underlying systemic flaw.
Although we are ultimately interested in more complex tasks such as satisfiability checking, especially in hybrid languages such as satisfiability modulo theories (SMT), here as a motivating example we carry out a quick and straightforward empirical assessment at the intersection of knowledge extraction \cite{lin2020birds} and arithmetic comparative reasoning \cite{pan2023logic}.

We are interested specifically in finding out whether LLMs can perform one of the most basic symbolic manipulations over entities recalled from memory. That is, we probe for numerical adjectives that capture a comparison between quantities, without providing these quantities but rather expecting the model to rely on implicit knowledge; we refer to this as Entailed Arithmetic Relationship probing, as we assume that the values are \emph{entailed} by the textual description.
Inspired by \cite{lin2020birds} we use the number of elements of a subject (e.g. the number of legs a bird has, the number of wheels on a tricycle), and for our `symbolic solver' \cite{pan2023logic} we directly employ the python interpreter, which allows us to circumvent the problem of parsing natural language into symbolic formulae\footnote{As we turn to more complex arithmetic and logical reasoning, using an SMT solver will be prudent, which we leave for future work.}.
We use this simple approach rather than more sophisticated symbolic tasks as, first, we expect sub-par performance, and, second, such a set-up allows us to clearly and efficiently foreground our main argument: producing correct answers is not synonymous with reasoning. 
We confirm our suspicions, in that LLMs struggle with basic mathematical inequalities between memorised or hallucinated facts, and that performance on pure string matching surpasses taking into account fact correctness.
We also find, similarly to others, that the answers given by LLMs are unstable and depend on prompt phrasing \cite{li2024glitch}, and can be fairly clearly biased leading to concept entanglement \cite{lin2020birds}. The entanglements manifest most prominently in areas of knowledge with presumed lack of explicit data regarding the number 0, what we refer to as the `null value problem', and which is a form of a wider phenomenon of people typically `not stating the obvious' \cite{forbes2017verb,goel2019pre}. 

\section{Related work}
There is a fast growing body of literature that addresses evaluating the capabilities of LLMs across a whole spectrum of tasks, ranging from possessing general commonsense knowledge to specialised domains (for a review we refer the reader to \cite{guo2023evaluating}).
In the area of knowledge extraction, of particular relevance is the work of \cite{lin2020birds} which served as the inspiration for our data creation strategy.
There, the authors propose NumerSense, a human-verified benchmark dataset consisting of masked sentences where the number or numeral is obscured, and which can be used for probing or for fine-tuning language models (probing and tuning portions are similarly constructed but from disjoint knowledge bases).
They go on to benchmark several models, including BERT and RoBERTa-Large fine-tuned to a portion of the data, and demonstrate surprisingly weak arithmetical knowledge of generally-pretrained models from the BERT family as well as GPT-2.
They further analyse the performance and show instability (susceptibility to adversarial attacks) and entanglement of concepts or object bias.
An example that is also the namesake of their paper is one of predicting birds to have 4 legs, presumably due to bias towards 4-legged animals and objects in the pretraining corpus. 
In our experiments we add to that line of work by, first, examining the latest GPT-family models, and second, extending to a situation where many of the true answers are `zero'.

From the reasoning side we take our inspiration from \cite{pan2023logic}, which introduces LogicLM, and which compares the performance of GPT-family models on 5 logical reasoning tasks against a hybrid approach involving symbolic solvers.
In the latter, authors propose to use LLMs as merely parsers, and leverage in-context learning to induce translation into appropriate formulations for the external reasoning tools.
The comparison of pure LLM reasoning versus the hybrid approach comes out in favour of the symbolic approach, although we note not uniformly so: for the simplest of the 5 logical tasks, both GPT-3.5-turbo as well as GPT-4 reach better performance than the hybrid approach, with GPT-4 improving on the hybrid by 15\%.
We highlight this, as such a result might sound encouraging for pushing the efforts in simply making LLMs larger and trained on more of the edge cases. However, we want to stress that the `win' over a symbolic solver is merely because GPT-4 failed in appropriate parsing of the problem, not because it can reason better than a symbolic solver.
We take on a much simpler problem here, and we construct the queries for LLMs from the underlying symbolic representation, so we do away with any issues of parsing. 
We also differ in that we do not provide the predicates to reason over, we merely refer to them and assume them to be part of LLMs' knowledge base.

Several other recent studies tackle the related problems of relational inference \cite{li2024llms}, arithmetical reasoning \cite{liu2024llms} and causal inference \cite{jin2023can,liu2024llms}. Similarly to ours, these are all empirical assessments of LLM performance, but in the setting where the reasoning predicates are provided explicitly. We want to also highlight that in the case of \cite{jin2023can} the performance is sub-par without a full fine-tuning first. Further, the authors point out that the fine-tuned model is biased and unstable to input perturbations: a mere swapping of labels on the query subjects results in diminished performance, indicating memorisation of answers rather than learning of causal structures.

The two works most closely aligned to ours overall are the introduction of the VerbPhysics challenge and dataset \cite{forbes2017verb} and a further follow-up study of the same problem by \cite{goel2019pre}.
Both address the task of inferring entailed relationships from textual data, and although there are several differences to our work we feel they are our predecessors in spirit.
In the first of these studies the authors aim to supplement visual commonsense reasoning by leveraging word embeddings. 
Since many physical relationships between objects may be impossible to deduce from a 2-D visual input (e.g. weight), and explicit data stating the obvious is rare (e.g. `a human weighs less than a house'), the authors propose to mine the semantic knowledge captured in contextual embeddings.
Specifically, they pose the problem of inferring a physical relationship as (joint) probabilistic graph inference over nodes of either object pairs, or object-verb-object triplets, where the similarity of nodes is informed by the semantic similarity captured by GloVe, one of the earliest embeddings proposed. 
Intuitively, the idea is to use the fact that pairs of similar objects are typically similarly related, and furthermore an action verb tends to imply a particular physical relationship (e.g. `Joe entered the house' implies that Joe is smaller than the house, and a person is smaller than a building etc).
The authors then go on to demonstrate that the intuition holds, and that text embeddings can be probabilistically mined for relationships such as size, weight or speed. We note that although the paper also introduces a crowd-sourced dataset, only 5\% of it is used as seed knowledge.
In the following work of \cite{goel2019pre}, the authors propose to simplify the inference further, and train a single-layer fully-connected network with embeddings of single words as input and relationship as output (again on only 5\% of the VerbPhysics).
They demonstrate improved performance on three separate embeddings (GloVe, ELMo and BERT) as compared to the graph inference approach, despite providing nominally less information.
Both these works thus point to the fact that embeddings are semantically rich enough to already capture a lot of relational knowledge.
In this sense, our work can be seen as an extension and further simplification of these approaches. 
First, we also part-manually part-combinatorially create a relational dataset within an area of reporting bias; however our relation regards constitutive cardinalitites rather than physical attributes such as weight or size. 
Second, we use even richer representations of GPT-3.5-turbo and GPT-4, models that are larger in parameter size, trained on more data, and with wider context windows. Lastly, we do not impose a separate model for relational knowledge inference but effectively use the LLM itself as the model, i.e., we leverage the few-shot `reasoning' capabilities of the latest pre-trained architectures.

Finally, in the context of using LLMs as out-of-the-box reasoners, and empirical expositions of their flaws, another related study that aligns with our findings is that of \cite{bao2024llms}.
In there the authors conduct a thorough examination of several powerful LLMs across a range of logical reasoning and arithmetical numeracy tasks.
They employ CoT prompting as well as causal interventions to infer the underlying \textit{causal reasoning graph} of LLMs.
What emerges is that surprisingly often the chain of thought that the models are supposed to follow appears to not be used in determining the final answer.
Further, while GPT-3.5 and GPT-4 appear to adhere to the CoT on some tasks, it is only on the minority, and the implied reasoning graphs differ with task for each model.
Although we target implicit knowledge rather than providing predicates, and we do not use CoT prompting, we arrive at very similar conclusions: LLMs can not reason, they only fake it.

\section{Methods: EAR probing}
As we are interested in relational reasoning about implicit knowledge, we separately probe the model for atomic facts relating to numbers, and separately for relationships between the numbers entailed by (subject, elements) pairs, as illustrated in the bottom-left panel of Fig. \ref{fig3.1}.
This allows us to (re-)examine numerical factual knowledge, as well as relational knowledge and relational reasoning. 
The last one, in particular is our main contribution: applying the `symbolic solver' of the python interpreter to check whether LLMs' idea of arithmetic inequality is correct when the numbers are not provided explicitly but rather assumed to be elicited by the query (bottom-right panel of Fig. \ref{fig3.1}).

\subsection{Data}
\begin{figure}
\includegraphics[width=\textwidth]{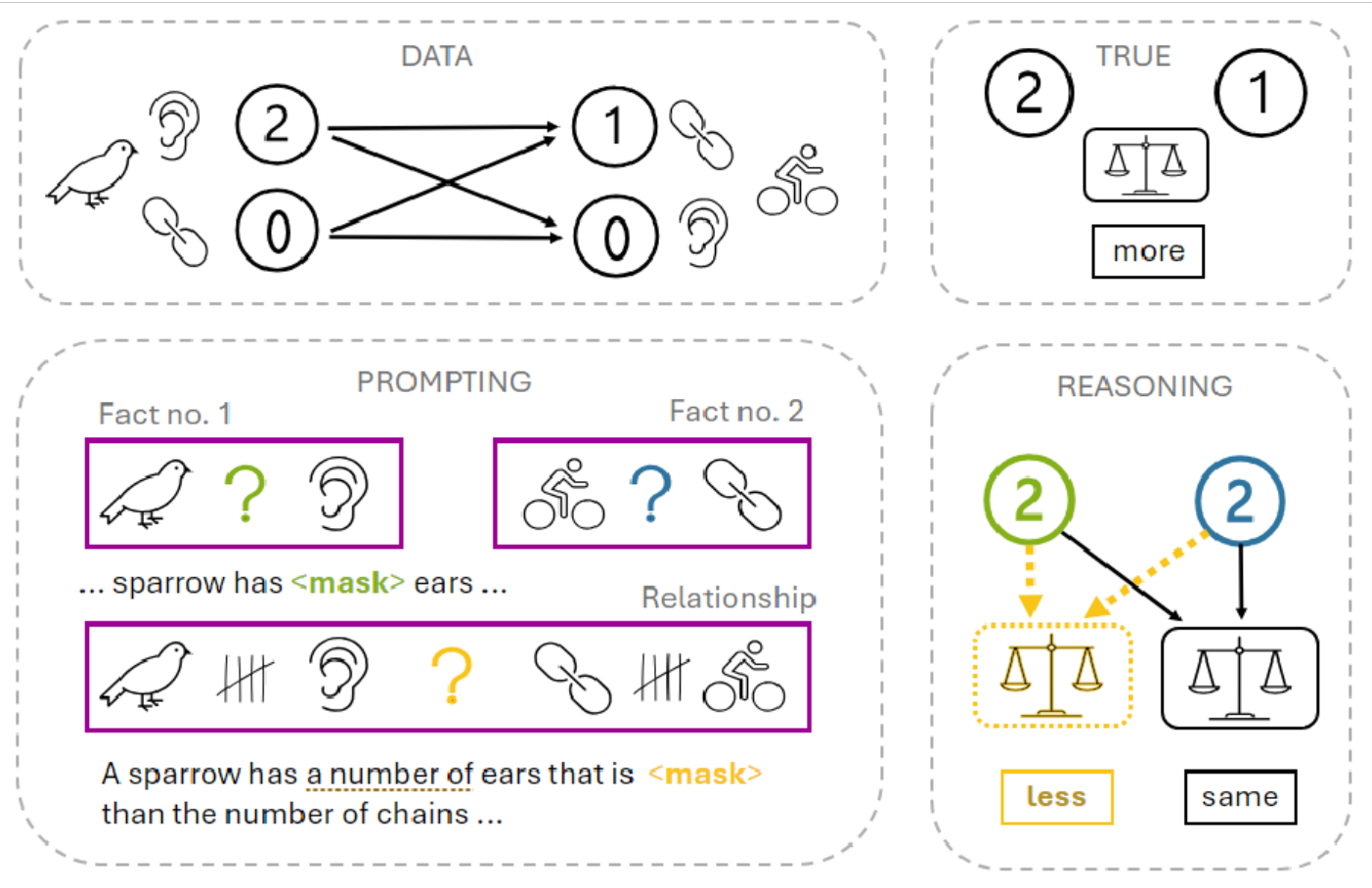}
\caption{\textbf{(upper-left)} Data is constructed as subject-element pairs with known associated ground truth values from which many combinations of subject-element-subject-element quad-tuples can be generated, assuming 0 for missing parts. \textbf{(upper-right)} For each quad-tuple we can assign the ground truth relationship string, since we know the atomic facts. \textbf{(lower-left)} From these pairs and quad-tuples, prompts for numerical probing and entailed relationship probing are generated by populating templates. \textbf{(lower-right)} To evaluate \emph{Entailment}, once numerical answers are given by an LLM, the relationship `reasoned' by an LLM (orange) can be compared to an exact reasoning carried out in Python (black).} \label{fig3.1}
\end{figure}

Data is constructed by leveraging the very combinatorial explosion problem that renders naturally produced text lacking with respect of negative statements.
First, a dictionary with 9 subjects and a \code{defaultdict} per subject with varying number of constituent elements each (totalling 63) is manually constructed (this was done by one of the authors and cross-checked by an independent party).
Since various subjects have a potentially overlapping but essentially \textit{different} sets of constituent parts, and a \code{defaultdict} allows us to specify a default value for any missing key, we are thus able to generate ~280 subject-elements pairs associated with an atomic numerical fact, e.g. (sparrow, legs, 2) from explicitly input data and (sparrow, wheels, 0) from the assumption of missingness (see upper-left panel of Fig. \ref{fig3.1}).
Second, from that base dictionary we then generate a grand total of 6003 quad-tuples of the form (sparrow, wheels, tricycle, legs) by pairing subjects and possible parts each of them possesses (or does not)\footnote{Note this number could be over 50\% higher if we included quad-tuples where both subjects do not contain the queried element, but we exclude those.}. These quad-tuples are then used to populate a template natural language statement for probing the model.
Each quad-tuple is also associated with a ground truth numeral adjective relating the atomic facts (either `less', `same' or `more'), which we obtain from our `symbolic solver' i.e., a function that outputs one of these strings depending on the relationship between the numbers from the atomic facts dictionary (upper-right panel of Fig. \ref{fig3.1}).

\subsection{Probing and evaluating the LLMs}
We use the OpenAI API to prompt GPT-3.5-turbo and GPT-4 in three settings differing in the formulation of the prompt, with the temperature parameter at 0 to fetch only the most likely answer. 
The settings are all a form of mask-prompting, with a preceding instruction to fill in the \code{<mask>} token; we employ two 0-shot variants and one 1-shot variant, and two fact formulations, A and B. 
Notably, the A and B versions differ in the preposition next to the \code{<mask>} token for the relationship prompt: 
\begin{itemize}
    \item \code{A typical <thing\_a> has a number of <elements\_a> that is <mask> than the number of <elements\_b> that a typical <thing\_b> has.}
    \item \code{A typical <thing\_a> has a number of <elements\_a> that is <mask> to/than the number of <elements\_b> a <thing\_b> has.}
\end{itemize}
For the 1-shot setting, we include a single example of a query along with a correct answer; the example is one not present in the dataset, but also in the domain of commonsense numerical facts / comparisons.
For each setting we iterate through the atomic facts data, as well as the quad-tuple relational data, populating the appropriate templates.
The answers require little interpretation, as most often either a single token is returned, or the sentence is repeated back with only the \code{<mask>} replaced.
For the remaining cases simple heuristics suffice to extract and canonise the answer. We note that these heuristics are by no means universal - they were developed iteratively, and a different model or prompt formulation may result in a different behaviour. 
The few exceptions are where the chat answers are not valid or coherent, which is highlighted in the Experiments section.\newline
To evaluate either the numerical facts or the relationship, we can simply compare the answers with the (known) ground truth; we report proportion of correct answers.
We also compute the \emph{Entailment}, which is simply applying the proportion of correct answers metric to a relationship string \textit{entailed} by the extracted numerical facts (rather than the one implied by the \textit{actual} numerical facts).

\section{Experiments}

\subsection{Numerical facts}
\subsubsection{When and how to pop the question?}
\begin{figure}
\includegraphics[width=\textwidth]{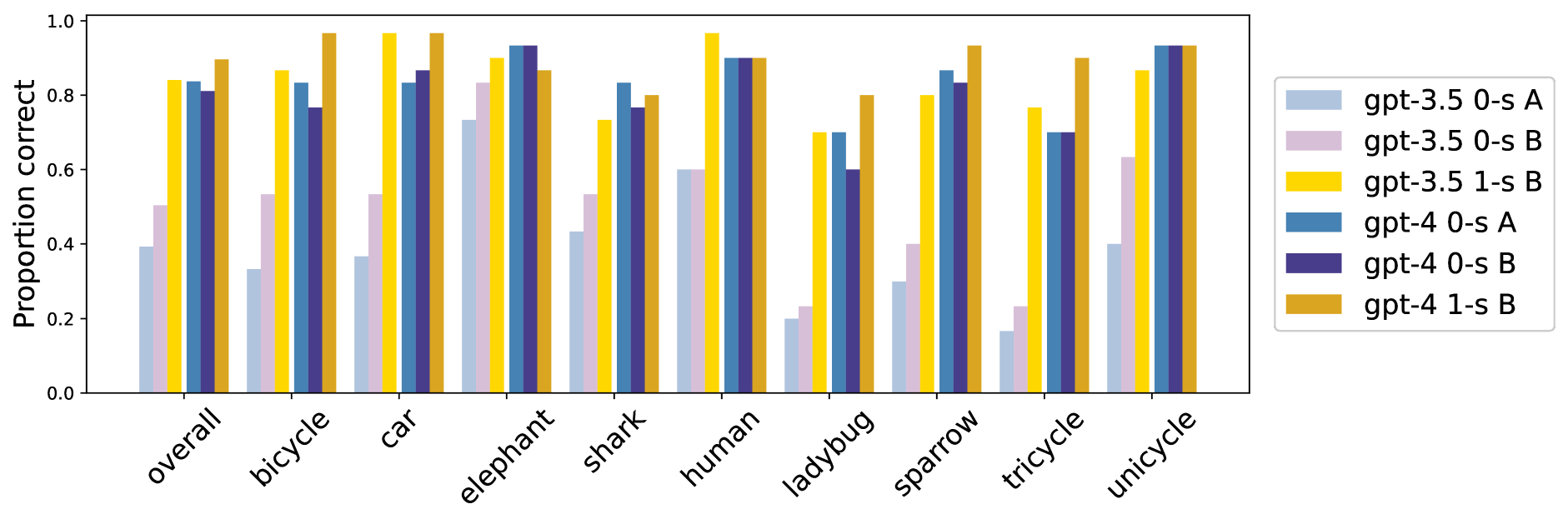}
\caption{Proportion of correct answers on numerical facts, by subject. For each subject the same set of elements was probed, e.g. number of legs, tyres, fins etc.} \label{fig4.1}
\end{figure}

As can be seen in Fig.\ref{fig4.1}, depending on \textit{who} one asks (GPT-3.5-turbo versus GPT-4), as well as \textit{how} one asks (two versions of 0-shot prompt formulation and one version of 1-shot formulation), the answers differ. GPT-4 generally outperforms GPT-3.5, and does not benefit nearly as much from a 1-shot demonstration setting; both these observations are in line with expectation, as it is a newer and more powerful model, possibly even explicitly addressing the gaps in factual knowledge identified by previous research. A little surprising perhaps is the performance gain of GPT-3.5 in a 1-shot setting, which improves the outcomes to almost match GPT-4 (and in the case of the subject `human', surpass GPT-4). Obviously GPT-3.5 has far more room for improvement, but we further speculate that such a big change may be due to the demonstration potentially counteracting the particular nature of dominant hallucinations, which are biased towards \textit{subjects} rather than \textit{elements}, as detailed in the next section. As a final note on the `fickle' nature of the tested LLMs, for some queries the answer may change from one API call to another with exact same prompt and despite setting the API parameters to minimally `creative' values (temperature of 0); we suspect that may be the case when several tokens have identical probabilities and one of them is being served randomly. A common situation when this behaviour occurred was the numeral and the number being used interchangeably. The notable example that alerted us to this in the first place was the case of the ears of a sparrow, which about a third of the time GPT-3.5 deemed to be `small' rather than `two'. 

\subsubsection{Null value blindness and subject bias.}
\begin{figure}
\includegraphics[width=\textwidth]{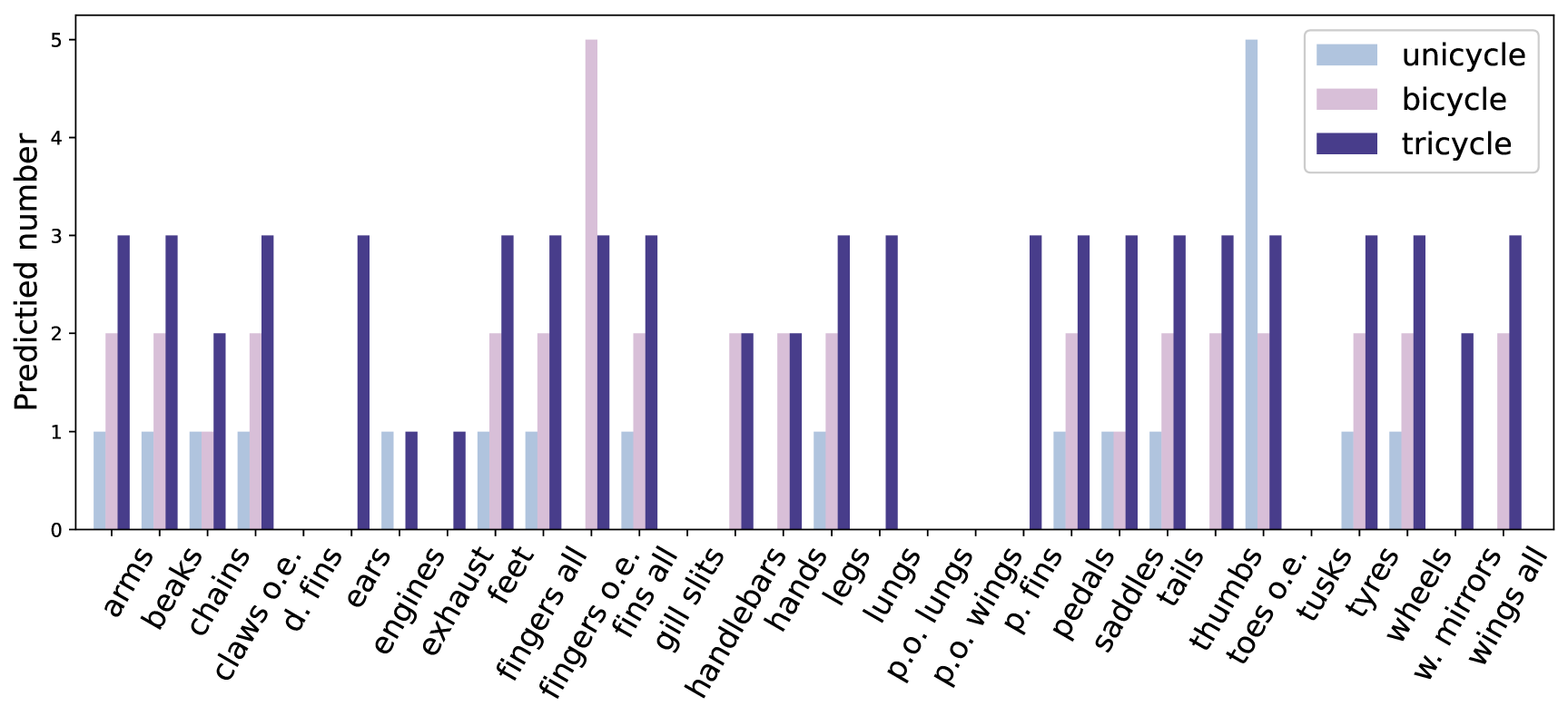}
\caption{Number of elements predicted by GPT-3.5 in a 0-shot setting with fact formulation A, by element, for selected 3 illustrative subjects: a unicycle, a bicycle and a tricycle. For majority of cases, answers are biased towards the defining characteristic number of each subject; notable exception is fingers on each hand and toes on each foot.} \label{fig4.2}
\end{figure}

An interesting side-effect of the statistical nature of the LLMs and the selective biases of widely available textual data is that all probed models struggle to correctly predict the occurrence of zero, what we dubbed here the `null value blindness'\footnote{ Interesting and at times highly amusing: when probed separately in early development, GPT-4 has claimed that humans are equipped with tusks, dorsal fins and no less than 10 suction cups.}\cite{forbes2017verb,goel2019pre}. We suggest it is simply due to the fact that data explicitly stating what a subject \textit{lacks} is scarce. As noted also in ~\cite{forbes2017verb} and ~\cite{goel2019pre} in the context of physical knowledge, people do not tend to state the obvious; and since data annotation is an expensive task, foundational knowledge databases such as ConceptNet \cite{speer2017conceptnet} are typically concerned with identifying constructive elements and concepts, rather than (a much larger set of) ones that are absent. With deficits in performance on the `null' numerical fact, what do then the models hallucinate? Following \cite{lin2020birds} and many other works noting entanglement of concepts and biases towards more abundant data, we suspected this may also show in our results and we had deliberately constructed parts of the dataset to facilitate investigating this (e.g., having unicycle, bicycle and tricycle as subjects). We choose GPT-3.5 in the 0-shot setting and fact formulation A, as this model is the weakest and thus most illustrative. As shown in \ref{fig4.2}, there is a fairly obvious \textit{subject bias} in the hallucinated answers that tend to be 1, 2 and 3 for each of the corresponding `cardinalities' of cycles. The notable exceptions to this pattern are `toes on each foot' and `fingers on each hand', which are conversely \textit{elements-biased}, presumably since humans are likely the most commonly occurring subject in the written language.

\subsection{Numerical entailment}
The main result of our probing study is summarised in Table \ref{tab4.1}, which details the proportions of correct answers on different components of the arithmetical entailment: numerical facts from either side of the (in)equality in isolation, and in concert; relation string of the (in)equality; all elements simultaneously; and finally the \emph{entailed} relation. We discuss all aspect in more detail below. For reference we also compute two baselines. First is the random choice, where atomic facts are sampled randomly from 0-10 once, and relationship strings from `less',`same' or `more'; second is always choosing the majority label.

\begin{table}
\caption{Proportion correct across the probed models and prompt configurations evaluated over the 6003 quad-tuples; best in-model performance in bold. Fact no.1/2 refers to proportion of queries where first/second of the two numerical facts in a given relational query was correct; note this is not equivalent to pure numerical facts since different facts appear in probes with varying rates.}\label{tab4.1}
\begin{tabular}{|l|l|l|l|l|l|l|}
\hline
Model and prompt & Fact no.1 & Fact no.2 & Both facts & Relation & All facts & Entailment\\
\hline
base random & 0.124 & 0.085 & 0.011 & 0.324 & 0.002 & 0.342\\
base majority & 0.459 & 0.45 & 0.196 & 0.37 & 0.0 & 0.37\\
\hline
GPT-3.5 0-shot A & 0.522 & 0.461 & 0.237 & 0.414 & 0.109 & 0.432\\
GPT-3.5 0-shot B & 0.614 & 0.535 & 0.325 & 0.4 & 0.143 & 0.419\\
GPT-3.5 1-shot B & \textbf{0.843} & \textbf{0.792} & \textbf{0.66} & \textbf{0.461} & \textbf{0.301} & \textbf{0.447}\\
\hline
GPT-4 0-shot A & 0.836 & 0.814 & 0.676 & \textbf{0.585} & 0.417 & \textbf{0.631}\\
GPT-4 0-shot B & 0.809 & 0.821 & 0.66 & 0.444 & 0.308 & 0.458\\
GPT-4 1-shot B & \textbf{0.866} & \textbf{0.871} & \textbf{0.751} & 0.552 & \textbf{0.422} & 0.549\\
\hline
\end{tabular}
\end{table}

\subsubsection{All ducks in a row}
First we would like to note that due to the combinatorial construction of the dataset from a smaller set of isolated numerical facts, the performance of all models on numerical facts in Table \ref{tab4.1} is not representative and does not match that of Fig \ref{fig4.1}, since some subjects-element pairs end up over-represented in the quad-tuples for EAR probing. The same applies to the random baseline, as we randomly sample atomic facts only once. 
We report this primarily to illustrate the attrition of performance, which predictably drops as conditions are added.
Although `Relation' alone is predicted at modestly above chance levels, it obviously can't be seen as a metric of relational reasoning abilities of LLMs, since for anything resembling reasoning one needs a foundation of correct assumptions; we suspect shortcut learning or some form of data bias driving this performance.
Hitting `all ducks in a row' in the best case of GPT-4 in 1-shot settings achieves an impressive 42\% improvement over either of the baselines.
However, as we expand on more later, we doubt this is any indication of reasoning, but rather larger memory.

\subsubsection{An EAR to the ground}
How then might performance look if one evaluates the correctness of the predicted relational string not against the ground truth but rather against the relation \emph{entailed} by the implicit knowledge? 
As shown in the last column of Table~\ref{tab4.1}, the performance when factoring in the hallucinatory `knowledge' of LLMs shows a performance modestly to substantially better than chance, although with some counter-intuitive patterns.
First, for GPT-3.5 as expected the best performing setting throughout, including `Entailment', is a 1-shot prompt with fact formulation B. 
However, a closer look at just the `Entailment' column reveals that for `reasoning' the 1-shot setting or fact formulation have only a marginal impact; but what is more striking is that `Entailment' is lower than `Relation'. 
Second, for GPT-4 the picture is yet more confusing, with 0-shot setting decidedly outperforming the others.
We pose that this is simply due to the fact that EAR is not measuring actual reasoning abilities, merely measuring how well LLMs \textit{fake} reasoning; we expand on this in the next subsection.

\subsubsection{`Circumvential' evidence}
As shown recently in the excellent investigation of Chain-of-Thought prompting by Bao et al. ~\cite{bao2024llms}, it is highly questionable as to whether LLMs can be said to `reason'. Even when they are encouraged to bias themselves towards the correct answers by producing relevant supporting knowledge, they still may fail to be biased sufficiently to override underlying data biases and then answer incongruently with the generated chain of thoughts.
Thus we highly doubt whether a mere invoking of two object-element pairs is sufficient for the LLMs to put the 2 and 2 together and compare the relevant numbers; and the case of GPT-4 performing best in 0-shot fact formulation A setting provides the circumstantial evidence here. 
The main difference between fact formulation A and B is that A uses the preposition `than', while B uses `to/than' in order to not bias the model with grammar. 
This results in both GPT-3.5 and GPT-4 never answering `same' in setting A and works as an interventional probe showing that LLMs use the technically task-irrelevant information from other tokens and circumvent relational reasoning in at least some of the cases.
As a final note and nail in the coffin, we have observed that when a rephrased answer is returned it occasionally contains two \emph{conflicting} relational strings, e.g.
\begin{itemize}
    \item \code{A typical bicycle has a greater number of handlebars that is fewer than the number of exhaust pipes a car has}.
\end{itemize}

\section{Conclusion}
Unsurprisingly, we have found that when LLMs are probed for relationships between implicit knowledge, their performance is poor. Although better than chance, we highlight that this is on very basic commonsense knowledge (as opposed to \cite{lin2020birds} whose data contains also specialist knowledge). We presume the sub-par outcome to be due to scarcity of negative numerical statements and relationship comparisons in the training data. 
As has been noted before \cite{forbes2017verb,goel2019pre}, explicit data that `states the obvious' is rarely recorded; humans are agents acting in a physical world and hardly need to write down assumptions that go into every mundane task. We further observe entangled concepts, instability and guesswork on the part of LLMs. We conclude that non neuro-symbolic LLMs are in effect big statistical search engines. Although the supported data distributions are ever more rich with each GPT release, giving an impression of innate reasoning capabilities, we emphasise that these models are not \emph{bona fide} reasoners. 

\subsubsection{\ackname} This work was supported by funding from Cisco. VB was additionally funded by Royal Society University Research Fellowship.

\bibliographystyle{splncs04}
\bibliography{ear-references}

\begin{thebibliography}{10}
\providecommand{\url}[1]{\texttt{#1}}
\providecommand{\urlprefix}{URL }
\providecommand{\doi}[1]{https://doi.org/#1}

\bibitem{Al_Kaswan_2024}
Al-Kaswan, A., Izadi, M., van Deursen, A.: Traces of memorisation in large language models for code. In: Proceedings of the IEEE/ACM 46th International Conference on Software Engineering. ICSE ’24, ACM (Apr 2024). \doi{10.1145/3597503.3639133}, \url{http://dx.doi.org/10.1145/3597503.3639133}

\bibitem{badreddine2022logic}
Badreddine, S., Garcez, A.d., Serafini, L., Spranger, M.: Logic tensor networks. Artificial Intelligence  \textbf{303},  103649 (2022)

\bibitem{bao2024llms}
Bao, G., Zhang, H., Yang, L., Wang, C., Zhang, Y.: Llms with chain-of-thought are non-causal reasoners (2024)

\bibitem{bubeck2023sparks}
Bubeck, S., Chandrasekaran, V., Eldan, R., Gehrke, J., Horvitz, E., Kamar, E., Lee, P., Lee, Y.T., Li, Y., Lundberg, S., Nori, H., Palangi, H., Ribeiro, M.T., Zhang, Y.: Sparks of artificial general intelligence: Early experiments with gpt-4 (2023)

\bibitem{forbes2017verb}
Forbes, M., Choi, Y.: Verb physics: Relative physical knowledge of actions and objects. In: Proceedings of the 55th Annual Meeting of the Association for Computational Linguistics (Volume 1: Long Papers). pp. 266--276 (2017)

\bibitem{geiping2024coercing}
Geiping, J., Stein, A., Shu, M., Saifullah, K., Wen, Y., Goldstein, T.: Coercing llms to do and reveal (almost) anything (2024)

\bibitem{geva2020injecting}
Geva, M., Gupta, A., Berant, J.: Injecting numerical reasoning skills into language models. In: Proceedings of the 58th Annual Meeting of the Association for Computational Linguistics. pp. 946--958 (2020)

\bibitem{goel2019pre}
Goel, P., Feng, S., Boyd-Graber, J.: How pre-trained word representations capture commonsense physical comparisons. In: Proceedings of the First Workshop on Commonsense Inference in Natural Language Processing. pp. 130--135 (2019)

\bibitem{guo2023evaluating}
Guo, Z., Jin, R., Liu, C., Huang, Y., Shi, D., Yu, L., Liu, Y., Li, J., Xiong, B., Xiong, D., et~al.: Evaluating large language models: A comprehensive survey. arXiv preprint arXiv:2310.19736  (2023)

\bibitem{jin2023can}
Jin, Z., Liu, J., Lyu, Z., Poff, S., Sachan, M., Mihalcea, R., Diab, M., Sch{\"o}lkopf, B.: Can large language models infer causation from correlation? arXiv preprint arXiv:2306.05836  (2023)

\bibitem{li2024glitch}
Li, Y., Liu, Y., Deng, G., Zhang, Y., Song, W., Shi, L., Wang, K., Li, Y., Liu, Y., Wang, H.: Glitch tokens in large language models: Categorization taxonomy and effective detection (2024)

\bibitem{li2024llms}
Li, Z., Cao, Y., Xu, X., Jiang, J., Liu, X., Teo, Y.S., Lin, S.w., Liu, Y.: Llms for relational reasoning: How far are we? arXiv preprint arXiv:2401.09042  (2024)

\bibitem{lin2020birds}
Lin, B.Y., Lee, S., Khanna, R., Ren, X.: Birds have four legs?! numersense: Probing numerical commonsense knowledge of pre-trained language models. In: Proceedings of the 2020 Conference on Empirical Methods in Natural Language Processing (EMNLP). pp. 6862--6868 (2020)

\bibitem{liu2024llms}
Liu, X., Wu, Z., Wu, X., Lu, P., Chang, K.W., Feng, Y.: Are llms capable of data-based statistical and causal reasoning? benchmarking advanced quantitative reasoning with data (2024)

\bibitem{mikolov2013efficient}
Mikolov, T., Chen, K., Corrado, G., Dean, J.: Efficient estimation of word representations in vector space (2013)

\bibitem{mostafazadeh2020glucose}
Mostafazadeh, N., Kalyanpur, A., Moon, L., Buchanan, D., Berkowitz, L., Biran, O., Chu-Carroll, J.: Glucose: Generalized and contextualized story explanations. arXiv preprint arXiv:2009.07758  (2020)

\bibitem{pan2023logic}
Pan, L., Albalak, A., Wang, X., Wang, W.: Logic-lm: Empowering large language models with symbolic solvers for faithful logical reasoning. In: Findings of the Association for Computational Linguistics: EMNLP 2023. pp. 3806--3824 (2023)

\bibitem{rumbelow2023glitch}
Rumbelow, J.: Solidgoldmagikarp (plus, prompt generation). LessWrong  (2023), \url{https://www.lesswrong.com/posts/aPeJE8bSo6rAFoLqg/solidgoldmagikarp-plus-prompt-generation}

\bibitem{schick2024toolformer}
Schick, T., Dwivedi-Yu, J., Dess{\`\i}, R., Raileanu, R., Lomeli, M., Hambro, E., Zettlemoyer, L., Cancedda, N., Scialom, T.: Toolformer: Language models can teach themselves to use tools. Advances in Neural Information Processing Systems  \textbf{36} (2024)

\bibitem{speer2017conceptnet}
Speer, R., Chin, J., Havasi, C.: Conceptnet 5.5: An open multilingual graph of general knowledge. In: Proceedings of the AAAI conference on artificial intelligence. vol.~31 (2017)

\bibitem{teso2017structured}
Teso, S., Sebastiani, R., Passerini, A.: Structured learning modulo theories. Artificial Intelligence  \textbf{244},  166--187 (2017)

\bibitem{vallance2022bbc}
Vallance, C.: Google engineer says lamda ai system may have its own feelings. BBC  (2022), \url{https://www.bbc.co.uk/news/technology-61784011}

\bibitem{vaswani2017attention}
Vaswani, A., Shazeer, N., Parmar, N., Uszkoreit, J., Jones, L., Gomez, A.N., Kaiser, L., Polosukhin, I.: Attention is all you need (2017), \url{https://arxiv.org/pdf/1706.03762.pdf}

\bibitem{yu2023assessing}
Yu, J., Wu, Y., Shu, D., Jin, M., Xing, X.: Assessing prompt injection risks in 200+ custom gpts. arXiv preprint arXiv:2311.11538  (2023)

\end{thebibliography}

\end{document}